\documentclass{soict}
\usepackage{multicol}

\title{Multifactorial Evolutionary Algorithm For Clustered Minimum Routing Cost Problem}

\author{Tran Ba Trung}
\affiliation{
  \institution{School of Information and Communication Technology, Hanoi University of Science and Technology}
  \city{Hanoi}
  \country{Vietnam}}
\email{batrung97@gmail.com}

\author{Huynh Thi Thanh Binh}
\affiliation{
  \institution{School of Information and Communication Technology, Hanoi University of Science and Technology}
  \city{Hanoi}
  \country{Vietnam}}
\email{binhht@soict.hust.edu.vn}

\author{Le Tien Thanh}
\affiliation{
	\institution{School of Information and Communication Technology, Hanoi University of Science and Technology}
	\city{Hanoi}
	\country{Vietnam}}
\email{thanhbok26b@gmail.com }

\author{Ly Trung Hieu}
\affiliation{
  \institution{School of Information and Communication Technology, Hanoi University of Science and Technology}
  \city{Hanoi}
  \country{Vietnam}}
\email{lytrunghieu5896@gmail.com}

\author{Pham Dinh Thanh}
\orcid{0000-0002-2550-9546}
\affiliation{
  \institution{The Faculty of Natural Science and Technology, Tay Bac University}
  \city{Sonla}
  \country{Vietnam}}
\email{thanhpd05@gmail.com}

\begin{abstract}
  \gls{clumrct} is applied in various fields in both theory and application. Because the \gls{clumrct} is NP-Hard, the approximate approaches are suitable to find the solution for this problem. Recently, \gls{mfea} has emerged as one of the most efficient approximation algorithms to deal with many different kinds of problems. Therefore, this paper studies to apply \gls{mfea} for solving \gls{clumrct} problems. In the proposed \gls{mfea}, we focus on crossover and mutation operators which create a valid solution of \gls{clumrct} problem in two levels: first level constructs spanning trees for graphs in clusters while the second level builds a spanning tree for connecting among clusters. To reduce the consuming resources, we will also introduce a new method of calculating the cost of \gls{clumrct} solution. The proposed algorithm is experimented on numerous types of datasets. The experimental results demonstrate the effectiveness of the proposed algorithm, partially on large instances. 
\end{abstract}



\copyrightyear{2019}
\acmYear{2019}
\setcopyright{acmcopyright}
\acmConference[SoICT 2019]{The Tenth International Symposium on Information and Communication Technology}{December 4--6, 2019}{Hanoi - Ha Long Bay, Viet Nam}
\acmBooktitle{The Tenth International Symposium on Information and Communication Technology (SoICT 2019), December 4--6, 2019, Hanoi - Ha Long Bay, Viet Nam}
\acmPrice{15.00}
\acmDOI{10.1145/3368926.3369712}
\acmISBN{978-1-4503-7245-9/19/12}

\begin{document}

\maketitle

\section{Introduction}
\glsresetall

The class of problems related to finding minimum cost clustered trees is among the most widely studied problems in applied mathematics and theoretical computer science. In general, clustered tree problems have many applications in the field of computer network design, computational biology, transportation and logistics as well as water resource management \cite{lin2017clumcrt,masone2019mrct}. Numerous problems with different cost function has received much attention such as Clustered Minimum Steiner Tree \cite{wu2015cluster_steiner_tree}, Clustered Shortest-Path Tree \cite{demidio2019hardness} and \gls{clumrct} \cite{lin2017clumcrt}.

Among those mentioned problems, \gls{clumrct} is one of the most newly investigated problems. This problem has been formally formulated in \cite{lin2017clumcrt}. Concretely, consider a connected,  undirected graph $G=(V, E, w)$ with nonnegative edge length function $w:E \rightarrow \mathbb{R}^{+}$. The vertices $V$ are partitioned into $k$ clusters $R = \{R_1,R_2,\ldots,R_k\}$. A spanning tree $T$ of $G$ is a clustered spanning tree if $T$ can be cut into $k$ subtrees by eliminating $k-1$ edges such that each subtree is a spanning tree for only one cluster. The \gls{clumrct} problem focuses on the routing cost, which is the sum of shortest path distance between any pairs of vertices given a clustered spanning tree $T$. The \gls{clumrct} problem is finding on graph $G$ a clustered spanning tree $T$ having the minimum routing cost. One practical realization of this problem is in computer network application where the communication terminals are vertices, and they are partitioned into many clusters. The communication between those terminals are restricted within a cluster and only a few terminals can be connected to another cluster for maximizing efficiency and other security concerns. Solving \gls{clumrct} is equivalent to facilitate the network architecture that consumes the minimum peer-to-peer communication resources.

The \gls{clumrct} has been proven to be NP-Hard by showing its relationship with an equivalent problem named clustered $st$-path problem, which is also proven to be NP-Hard \cite{lin2017clumcrt}. However, it is a pressing problem and approximate solutions are acceptable in practice. There are various approaches for finding approximate solutions for these type of intractable problem such as approximation algorithms, heuristic algorithms or meta-heuristic algorithms. However, methods to solve \gls{clumrct} are limited to only a 2-approximation algorithm for the case of three clusters \cite{lin2017clumcrt}. In the practical application of \gls{clumrct} for computer network, there would be multiple designs of the computer network topologies with various ways to set up the clusters constraint, thereby arising the problem of solving not only one \gls{clumrct} but multiple \gls{clumrct} instances. These instances do not exist in isolation; thus, effectively sharing the underlying pattern of minimum cost clustered tree for one instance may result in a better solution in another instance. 

Motivated by that practical indication, this paper aims to design a \gls{mfea} \cite{gupta2016mfea} to simultaneously solve multiple minimum cost clustered trees problem instances together. \gls{mfea} is a variant of \gls{ea}~\cite{pham2015effective,thanh2015new}, which is classified as a meta-heuristic algorithm that capable to provide a near-optimal solution for those mentioned hard problems in an acceptable timing limit. It is also being an algorithmic realization of a widely-study multitasking optimization scheme. It is attracting considerable interest because of its capability to optimize and facilitate better solutions for multiple related problems. The driving force behind \gls{mfea} is its genetic transfer using crossover across different tasks that share a partial proportion of high-quality solution from one task to another \cite{gupta2016genetic_transfer}. If problems optimized by \gls{mfea} having the highly correlated fitness landscape and near global optima, the high-quality solution from one task will definitely having high performance in another, leading to significant improvement in every constitutive task \cite{bali2017ldamfea,gupta2016landscape}. 

In particular, our contributions can be summarized as follow:
\begin{itemize}
  \item We proposed a novel evolutionary algorithm to solve the \gls{clumrct}, including a two levels meta-heuristic crossover and mutation operators on adjacency list representation of the problem.
  \item By following the generic procedure of \gls{mfea}, we expanded the representation and operators to facilitate a common platform for facilitating knowledge exchange between different instances of \gls{clumrct} of the various number of clusters and cluster size.
\end{itemize}

This paper is organized as follows. Section \ref{sec:problem} again reformulates the \gls{clumrct} and presents comprehensive details on the way to efficiently compute its cost function. Section \ref{sec:related_work} examines the literature of the considered problem and the development of \gls{mfea}. The next section provides the two proposed algorithms. An empirical study to analyze the effectiveness of the proposed approaches is provided in Section \ref{sec:result}. A discussion of the result and future direction is presented in Section \ref{sec:conclusion}

\section{Problem formulation}
\label{sec:problem}
 Given a weighted undirected graph  $G = (V, E, w)$. $V$ and $E$ are the vertex and the edge sets, respectively. In particular, all paths in the graph $G$ are simple path. A path in $G$ is simple if no vertex appears more than once on it. $w$ is a nonnegative edge length function $w:E \rightarrow \mathbb{R}^{+}$. An edge between vertices $u$ and $v$, where $u, v \in V$, is denoted by $(u, v)$. The weight of edge $(u, v)$ is denoted by $w(u, v)$. A graph $G$ is connected if all pairs of vertex within the graph is connected. A pair of vertices $(u, v)$ is connected if $G$ contains a path from u to v. For a graph $G$, $V(G)$ and $E(G)$ denote the vertex and the edge sets, respectively. For a vertex subset $U$, the subgraph of $G$ induced by $U$ is denoted by $G[U]$.


The authors define the H-graph ${G'} = ({V'}, {E'})$ where each vertex represents a cluster $R_i$. An edge connecting two vertices of $G’$ exists if the two corresponding clusters in G are connected by at least one edge. 
A spanning tree $T$ of the graph $G$ is a clustered spanning tree if all the vertices in the same cluster $R_i$ are clustered together in $T$. It means that $T$ can be cut into $k$ subtrees by removing $k-1$ edges such that each subtree is a spanning tree for one cluster $R_i$.

The objective of \gls{clumrct} problem mainly focuses on the routing cost $c(T)$. The routing cost $c(T)$ is the total distance over all pairs of vertices in the connecting path defined by the clustered spanning tree $T$:
\begin{equation}
  c(T) = \sum_{u,v \in V(T)} d_T(u, v)
\end{equation}
where $d_T(u, u)$ is the distance between $u$ and $v$ on $T$. The \gls{clumrct} problem searches for the clustered spanning tree $T$ that minimize the routing cost:
\begin{equation}
  \underset{T}{\text{minimize}} \; c(T) = \sum_{u,v \in V(T)} d_T(u, v)
\end{equation}

\section{Related work}
\label{sec:related_work}
In recent years, there has been considerable interest in solving the minimum cost clustered tree problem and its variants \cite{lin2017clumcrt}. Major developments have been carried out in developing approximate algorithms for solving the NP-hard \gls{clusteiner} variant and deriving the Steiner ratio of the problem \cite{wu2015cluster_steiner_tree}. \citeauthor{chen2017clustered_steiner_tree} subsequently proposed another approximate algorithm that improved the performance ratio by 40\% under the condition of distance metric satisfying the triangular inequality. Another NP-hard variant of the general class of minimum cost clustered tree problem called \gls{cluspt} has also been widely investigated in recent literature, under theoretical viewpoint \cite{demidio2019hardness} as well as developing approximation algorithms for this problem \cite{binh2019dijkstra, thanh2019greedy}. Following studies from \citeauthor{thanh2018cmfea} suggested another approach for solving this class problem more effectively by adopting the evolutionary multitasking paradigm \cite{thanh2018cmfea, binh2018emfea}.

\gls{clumrct} that we are focusing on this study is also a typical variant of the general minimum-cost clustered tree that having numerous practical applications. For instance, applications of \gls{clumrct} problem arise in the fields of network design, computational biology and transportation \cite{lin2017clumcrt}. \gls{clumrct} is proven NP-hard by a transformation to the equivalent problem of. Because of its NP-hardness, a popular method to tackling the \gls{clumrct} would be developing approximation algorithm, heuristic or meta-heuristic algorithm. However, to the best of our knowledge, there are no published study has proposed such method for this problem.

In recent literature, \gls{mfea}~\cite{bali2019multifactorial} is a newly founded meta-heuristic algorithm that has been gaining much attention due to its ability to solve multiple hard problems more effectively than the traditional \gls{ea} \cite{gupta2016mfea}. Some preliminary work was carried out suggesting that \gls{mfea} is being able to capitalize the knowledge overlap across relevant tasks to facilitate better objective value on solving multiple combinatorial problems concurrently \cite{zhou2016vrp, yuan2016tsp}. Other studies attempt to investigate the theoretical foundation of \gls{mfea} as well as improve this algorithm such that its performance would not be impeded by solving tasks that are two conflicting in their fitness landscape and global optima \cite{bali2018mfeaii,chen2019matde}. Also, there is no prior attempt to design a \gls{mfea} for solving \gls{clumrct}.

\section{Proposed algorithm}
\label{sec:propose}
In this section, we introduce \gls{mfea} to solve multiple \gls{clumrct} problems simultaneously. A \gls{clumrct} problem corresponding to a particular task in \gls{mfea}. Tasks are different number of clusters and different dimension is encoded into a unified individual. Each unified individual is divided into two-level, the first deal with constructing the spanning tree inside the clusters, the second problems deal with constructing tree connect between clusters. In the entire optimization process, we introduce novel two-levels genetic operators to construct spanning tree starting from the smallest task to the biggest task. As a result, the unified individual is the biggest tree contains the solution of all tasks.

\subsection{Individual encoding, Population Initialization and Decoding method}
An individual in the unified search space encompassing the genetic material of all of the tasks. An individual in the unified search space consists of the different solutions of \gls{clumrct} problems corresponding to tasks. An individual is a spanning tree which is the number of clusters is the maximum number of the clusters from all of the tasks, and the cluster inside takes the maximum number of vertices from the all of all tasks. The clusters in each instance of \gls{clumrct} problem is ascending sorted by the number of vertices before encoding into a unified individual to maximize the number of overlapping dimensions of different tasks. As a result, a unified individual captures solution for all task and the decoding method only finds the corresponding sub-graph with the task to produce valid solution for particular \gls{clumrct} task. After defining structures of an individual, we generate a new individual for the \gls{clumrct} problem. Initial individual is constructed randomly by applying PrimRST algorithms in the smallest task, all initial bigger tasks are generated from the smaller task and input graph by detecting and delete cycle until gaining spanning tree for the biggest task. The detail about encoding, decoding, and initializing individual are presented in \cite{binh2018emfea}.

\subsection{Crossover operator}
We introduce a new crossover operator for individuals in unified search space. In particular, the novel crossover operator generates two offsprings from two parents. Two offsprings are constructed from the set of edges of two parents. The crossover operator starts by selecting two individuals in unified search space from the current population to determine the common edges belong to each individual, the length of this set represents the similarity of two parents. If two individuals have more common edge meaning that two individual close similarities. The algorithm determines the set of edges the only belong to a particular parent. 
  
\setlength{\intextsep}{3pt}
\renewcommand{\scalefigure}{0.22}
\begin{figure*}
  \centering
  \begin{subfigure}[b]{.40\linewidth}
    \centering
    \includegraphics[scale=\scalefigure]{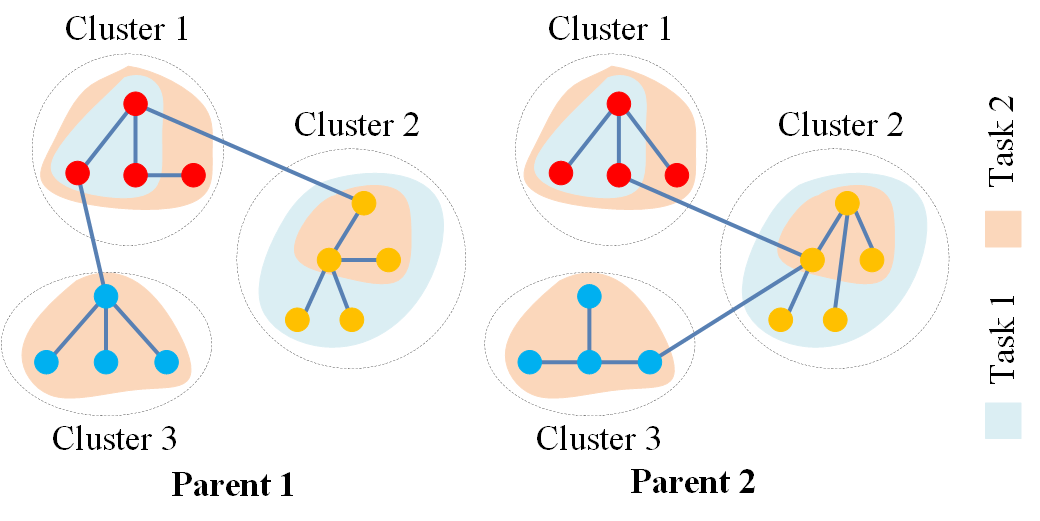}
    \caption{}
    \label{fig:Crossover_a}
  \end{subfigure}   
  \begin{subfigure}[b]{.24\linewidth}
    \centering
    \includegraphics[scale=\scalefigure]{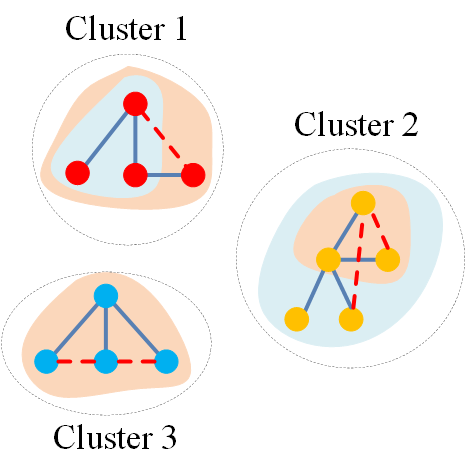}
    \caption{}
    \label{fig:Crossover_b}
  \end{subfigure}
  \begin{subfigure}[b]{.24\linewidth}
    \centering
    \includegraphics[scale=\scalefigure]{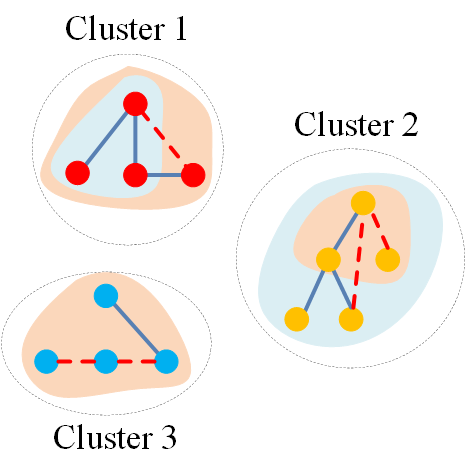}
    \caption{}
    \label{fig:Crossover_c}
  \end{subfigure}

  \renewcommand{\scalefigure}{0.22}
  \begin{subfigure}[b]{.23\linewidth}
  \centering
  \includegraphics[scale=\scalefigure]{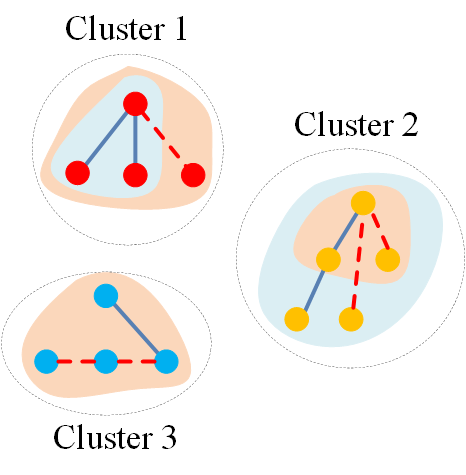}
  \caption{}
  \label{fig:Crossover_d}
  \end{subfigure}
    \renewcommand{\scalefigure}{0.20}
    \begin{subfigure}[b]{.20\linewidth}
    \centering
    \includegraphics[scale=\scalefigure]{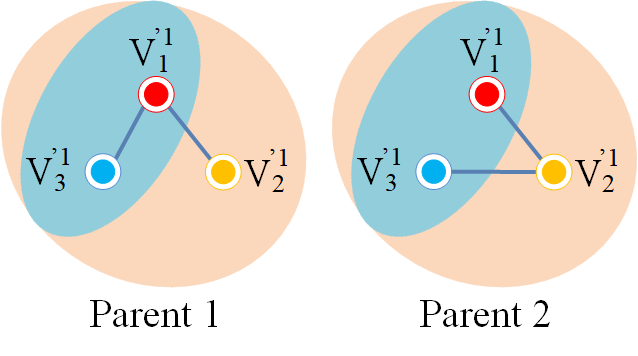}
    \caption{}
    \label{fig:Crossover_e}
  \end{subfigure}
    \begin{subfigure}[b]{.12\linewidth}
    \centering
    \includegraphics[scale=\scalefigure]{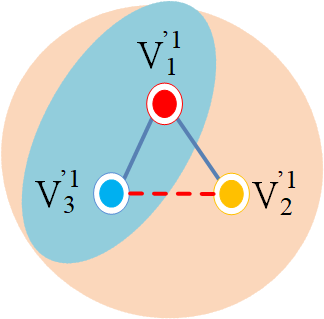}
    \caption{}
    \label{fig:Crossover_f}
  \end{subfigure}
    \begin{subfigure}[b]{.12\linewidth}
    \centering
    \includegraphics[scale=\scalefigure]{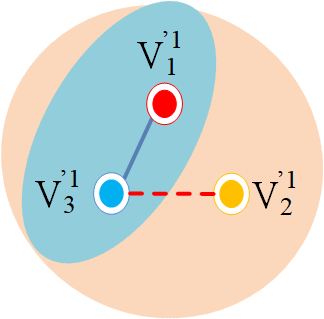}
    \caption{}
    \label{fig:Crossover_g}
  \end{subfigure}
    \begin{subfigure}[b]{.22\linewidth}
    \centering
    \includegraphics[scale=\scalefigure]{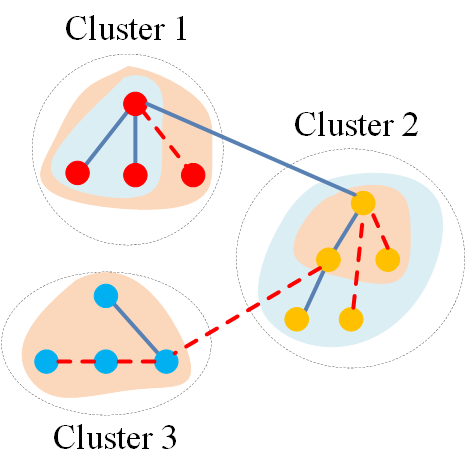}
    \caption{}
    \label{fig:Crossover_h}
  \end{subfigure}
  \caption{An example of crossover process applying two parents for MFEA with two tasks}
  \label{fig:Crossover}
\end{figure*}

After that, a new offspring is constructed based on the combination of a set of common edges and a set of edges only belong to its parent. The crossover operator is process in two-levels: the first is applying for spanning inside cluster and the seconds one is applying for spanning tree which is connect among clusters. The details as presenting in Algorithm~\ref{alg:Proposed_crossover_operator}.

\begin{algorithm}[tbp]
  \KwIn{Two parents: $T_{t}=\left(V,E_{t}, C\right), t = 1,2;$}
  \KwOut{Two offspring $S_1=(V_s, E_{s_1})$ and $S_2=(V_s, E_{s_2});$ }
  \BlankLine
  \Begin
  { 
    $V_s \leftarrow V$; \quad $\theta \gets |C|$ \;
    \lForEach{cluster $j^{th}$}
    {   
      $S^{j}_{1}, S^{j}_2 \leftarrow DSTX(T_1 ,T_2)$
    }
    
    Determine two H-Graphs $G'_1$ and $G'_2$ from $T_1$ and $T_2$, respectively\;
    
    $T'_1, T'_2 \leftarrow DSTX (G'_1, G'_2)$\; 
    $E_{s_1} \leftarrow (\cup_{j=1}^{\theta} E(S^{j}_1)) \cup E(T'_1)$;
    $E_{s_2} \leftarrow (\cup_{j=1}^{\theta} E(S^{j}_2)) \cup E(T'_2)$\;
    \Return $S_1, S_2$
  }
  \caption{Proposed crossover operator}
  \label{alg:Proposed_crossover_operator}
\end{algorithm}

The steps of the DSTX algorithm is presented in Algorithm~\ref{alg:DSTX}. DSTX generates two spanning trees $S_1, S_2$. Line 2, 3 define two sets of edges only belong to a particular parent. Line 8 $\rightarrow$ 23 generates two offsprings spanning tree from the smallest task to the biggest task. Line 9 $\rightarrow$ 15 to generate offspring $S_1$. In each particular task, the algorithm detects the cycles which are created by adding new edges to parents solution, then delete an edge in each cycle path to obtain the new solution. Removing an edge will be created a new valid solution for the current task but without disrupting the solution for the smaller task.

\begin{algorithm}[htbp]
  \KwIn{Two spanning trees $T_1=(V,E_1)$ and $T_2= (V, E_2)$ where $V = V_h \supseteq V_{h-1} \supseteq \ldots \supseteq V_1 $}
  \KwOut{Two spanning tree $S_1=(V, E_{s_1})$ and  $S_2 = (V, E_{s_2})$}
  \BlankLine
  \Begin
  {     
    $E'_1 \leftarrow E_1\backslash E_2$; \qquad
    $E'_2 \leftarrow E_2\backslash E_1$; \qquad
    $ t \leftarrow  |E'_1|$\; 
    $ k \leftarrow $ Generate a random number $k$ from $[1,t]$ \;
    $E_{s_1} \leftarrow$ Add $k$ random edges from $E'_1$ to $T_2$ \; 
    $E_{s_2} \leftarrow$ Add $k$ random edges from $E'_2$ to $T_1$ \; 
    \For{$i\leftarrow 1$ \KwTo $h$}
    {
        \ForEach{ edge $e = (v_1, v_l)$ in $E'_1$}{
        \If{${v_1 ,v_l} \in V_i$}{
          Find the path   $P = (v_1,v_2,\ldots,v_l)$ between $v_1$ and $v_l$ in the set of edges $E(T_1) - \{e\}$ \;
          Choose a vertex $v_j \in P$ such that $v_j \in V_{i} \backslash V_{i-1}$\;
          Delete edge $(v_j,v_{j+1})$ from $S_1$ \;
        }
      }
    \ForEach{ edge $e = (v_1, v_l)$ in $E'_2$}{
      \If{${v_1 ,v_l} \in V_i$}{
       Find the path  $P = (v_1,v_2,\ldots,v_l)$ between $v_1$ and $v_l$ in the set of edges $E(T_2) - \{e\}$ \;
        Choose a vertex $v_j \in P$ such that $v_j \in V_{i} \backslash V_{i-1}$\;
        Delete edge $(v_j,v_{j+1})$ from $S_2$ \;
      }
    } 
      
  }   
    \Return $S_1, S_2$ \;
  }
  \caption{DSTX} 
  \label{alg:DSTX}
\end{algorithm}
The Figure \ref{fig:Crossover} illustrates the crossover operator on two parents. The Figure~\ref{fig:Crossover}(b-d)  presents crossover for each cluster, other figures present crossover steps for H-Graphs. Figure~\ref{fig:Crossover_b} depicts the sub-graph when combine the sub-graph on clusters in Parent 1 and the set of edges $E'_2$ only belong to Parent 2.  The red dash edges in Figure~\ref{fig:Crossover_d} belong to parent 2 obtained by performing the DSTX. Figure\ref{fig:Crossover_e} illustrates the sub-graph after building the spanning tree for clusters. Figure \ref{fig:Crossover_d} and Figure~\ref{fig:Crossover_g} show graph after combining H-Graphs in second stages. The offspring obtained by crossover operator is depicted in Figure~\ref{fig:Crossover_h} after combining two graph in Figure~\ref{fig:Crossover_d} and Figure~\ref{fig:Crossover_g}.

 \subsection{Mutation operator}
   This mutation operator to generate new individual based on applying small variation on the original individual. The main idea of the proposed mutation operator (PMO) is that the PMO adds an edge in the input graph G but not in the individual to create a cycle and then removes an edge in the cycle to create a new spanning tree. The details of the MPO is presented in Algorithm~\ref{alg:ProposedMutationOperator}.
    
  \setlength{\intextsep}{0pt}
  \begin{algorithm}[htbp]
  	\KwIn{Input graph $G = (V_G, E_G, C_G)$ where $C_G = C^1 \cup C^2 \cup \ldots \cup C^{\theta}, C^p \cap C^q = \emptyset, \ \forall p \neq q$; An individual $T=(V,E,C)$}
  	\KwOut{A Tree $T'= (V', E')$}
  	\BlankLine
  	\Begin
  	{   
  		$T' \leftarrow T$; \qquad 
  		Select a random cluster $C^j$\;  
  		$e=(u,v) \gets$ Select randomly an edge in $E(G[C^j])$ but not in  $E(T'[C^j])$\;
  		$P_{T'}(u,v) \gets$ Path between vertex $u$ and vertex $v$ in $T'$\;
  		$e' \gets$ Select randomly an edge in $P_{T'}(u,v)$\;
  		$E(T') \gets (E(T) - \{e\}) \cup \{e'\}$ \Comment{exchange edges $e$ and $e'$}\;
  		\Return $T'$;	
  	}
  	\caption{Proposed mutation operator}
  	\label{alg:ProposedMutationOperator}
  \end{algorithm}
  \setlength{\intextsep}{0pt}

\subsection{Evaluation method}
In this sub-section, we present a method to calculate the cost of \gls{clumrct} solution in $O(n)$ time complexity. The cost function of \gls{clumrct} is the sum of connectivity cost between all pairs of vertices in the graph. The solution of \gls{clumrct} is spanning tree, thus, traveling all vertices in $O(n)$ time complexity and removing an edge from spanning tree will create two connected components. Given $T=(E,V)$ is the \gls{clumrct} solution, the cost of two adjacency vertices on solution $T$ are calculated as the following formula  $C(v_1, v_2) = |V1|*|V2|*w(v_1, v_2)$ where $|V1|$ is the number of vertices in connected component  contains $v_1$ after remove  $e = (v_1, v_2)$ from $T$. $|V2|$ is the number of vertices in connected component contains $v_2$ after remove  $e = (v_1, v_2)$ from $T$. $w(v_1, v_2)$ is connecting cost between two adjacency vertices $v_1$ and $v_2$.

The cost of \gls{clumrct} solution $T$ is calculated as following four-step procedure: 
\begin{itemize}
	\item \textbf{Step 1:} Assign a random vertex as start vertex $s$.
	\item \textbf{Step 2:} BFS algorithms start with $s$ vertex then store the order of visiting to the visited list.
	\item \textbf{Step 3:} The cost of two adjacency veritices on solution $T$ are calculated as following: Travel bottom-up (leaves to root) in visited list to update the label of current vertex to $ l = d+ 1 $, where $d$ is the number of descendant vertices of current vertex. Update the label of ancestor vertex to $l_a = l_a +l$. The cost between current vertex $v_1$ and its ancestor $v_2$ is calculated by formula $C(v_1, v_2) = l* (n- l)*w(v_1, v_2)$. Where, $l$ is the current vertex label, $n$ is the number of vertices, $w(v_1, v_2)$ is connecting cost between current vertex $v_1$ and its ancestor $v_2$.
	\item \textbf{Step 4:} The cost function of \gls{clumrct} is calculated follow as: $\sum_{v_1 \in V} \sum_{v_2 \in V} C(v_1, v_2)$. Where, $C(v_1,v_2)$ is the cost of two adjacency $v_1$ and $v_2$ in \gls{clumrct} solution $T$.
\end{itemize}

\section{Experimental results}
\label{sec:result}
\subsection{Problem instances}
To the best of our knowledge, there is no publicly available set of benchmark for the \gls{clumrct} problem. Therefore, we utilized a set public test instances for an equivalent clustered tree problem which is the \gls{cluspt} dataset version 3 on Mendeley \cite{thanh2018cluspt_instance}. 
This dataset included six distinct types of instances that were generated by various algorithms \cite{binh2019dijkstra} and categorized into two types conformable to their dimension. The instances were appropriate for evaluating cluster problems \cite{mestria2012grasp}.  However, to test the effectiveness of proposed algorithms to solving the \gls{clumrct} instead of \gls{cluspt}, we ignore the information of the global source vertex of this \gls{cluspt} dataset. For evaluation of the proposed algorithms, instances with dimensionality from 30 to 500 were selected.

\subsection{Experimental setup}
To evaluate the performance of the proposed algorithm for the \gls{clumrct}, the authors implemented the metaheuristic algorithm \gls{ea} and \gls{mfea}. Then, the results of two algorithms for each instance were compared with each other in term of the best and average fitness of the optimized solutions.

Each scenario was simulated for 30 times on the computer (Intel Core i7 – 4790, 16GB RAM), with a population size of 100 individuals evolving through 500 generations. The random mating probability is 0.5 and the mutation rate is 0.05. The source codes were installed by Java programming language.

\subsection{Experimental criteria}
Criteria for evaluating the quality of the result of the algorithms are presented in the following table:
\begin{center}
    \label{tab:Criteria}
    \begin{tabular}{p{2.5cm} p{5.5cm}}
        \hline
        \textbf{Criteria} & \textbf{Description} \\ 
		\hline
        Average (Avg)     & Average function value over all runs \\ 
        Best-found (BF) & Best function value achieved over all runs \\ 
        \gls{rpd}  & The diffrence between the average costs of two algorithms\\ 
        \hline
    \end{tabular} 
\end{center}

In order to compare the quality of the \gls{clumrct} solutions received from \gls{ea} and \gls{mfea}, \gls{rpd} is used to compute the difference between the average costs of two algorithms. The \gls{rpd} is calculated by the following formula:
$RPD(A,B)= \dfrac{C_B-C_A}{C_B}*100\%$
where $C_B$ is the average cost of a solution obtained from \gls{ea}, $C_A$ is the average cost of a solution obtained from \gls{mfea}.

\subsection{Experimental results}
The experimental results show that the quality of the \gls{clumrct} solution of \gls{mfea} algorithm is better than \gls{ea} algorithm in most scenario. Particularly, MEFA’s results are better than EA’s results in all instances in Type 3, Type 4, Type 1 Large, Type 5 Large and Type 6 Large. It means that the larger instance's size, the more \gls{mfea} tends to outperform than \gls{ea}. There is a noteworthy point that with large instances, the results obtained by \gls{mfea} are always better than one obtained by \gls{ea} in both average result and best-found in any Type. The comparison of results obtained by \gls{ea} and results obtained by the proposed algorithm is presented in detail in the Table~\ref{tab:result:summary}.

\setlength{\intextsep}{3pt}
\begin{table}[htbp]
  \centering
  \caption{The summarized results demonstrate the proportion of instances that \gls{mfea} outperformed \gls{ea} in term of number of instances having higher best-found fitness and average fitness}
  \setlength{\tabcolsep}{2pt}
  \begin{tabular}{c c c c c c c c c}  
    \hline
     & \multicolumn{2}{c}{\textbf{Type 1}} & \textbf{Type 3} & \textbf{Type 4}  & \multicolumn{2}{c}{\textbf{Type 5}} & \multicolumn{2}{c}{\textbf{Type 6}} \\
     \cline{2-9}
                      & Small       & Large       & Large & Large   & Small       & Large       & Small       & Large     \\ 
    \hline
    BF        & 22          & 10          & 5     & 8       & 15          & 14          & 33          & 13        \\
    Avg           & 20          & 10          & 5     & 8       & 15          & 14          & 33          & 13        \\
    Total             & 24          & 10          & 5     & 8       & 16          & 14          & 34          & 13
  \end{tabular}
  \label{tab:result:summary}
\end{table}

The details results obtained by algorithms are presented in Table~\ref{tab:result:type34} -- Table~\ref{tab:result:type1}.
In this table, the italic red cells on a column to show that on those instances, this algorithm outperforms than the other algorithm. The biggest RPD(MFEA, EA) is 98.59\% while the biggest RPD(EA, MFEA) is 16.9. 


\subsection{Convergence trend}
The functions in~\cite{gupta2016mfea} was used for computing the normalized and averaged normalized objectives for analyzing the proposed MFEA algorithm’s convergence trends.

\renewcommand{\scalefigure}{0.6}
\begin{figure*}[htbp]
	\centering
	\begin{subfigure}[b]{.32\linewidth}
		\centering
		\includegraphics[scale=\scalefigure]{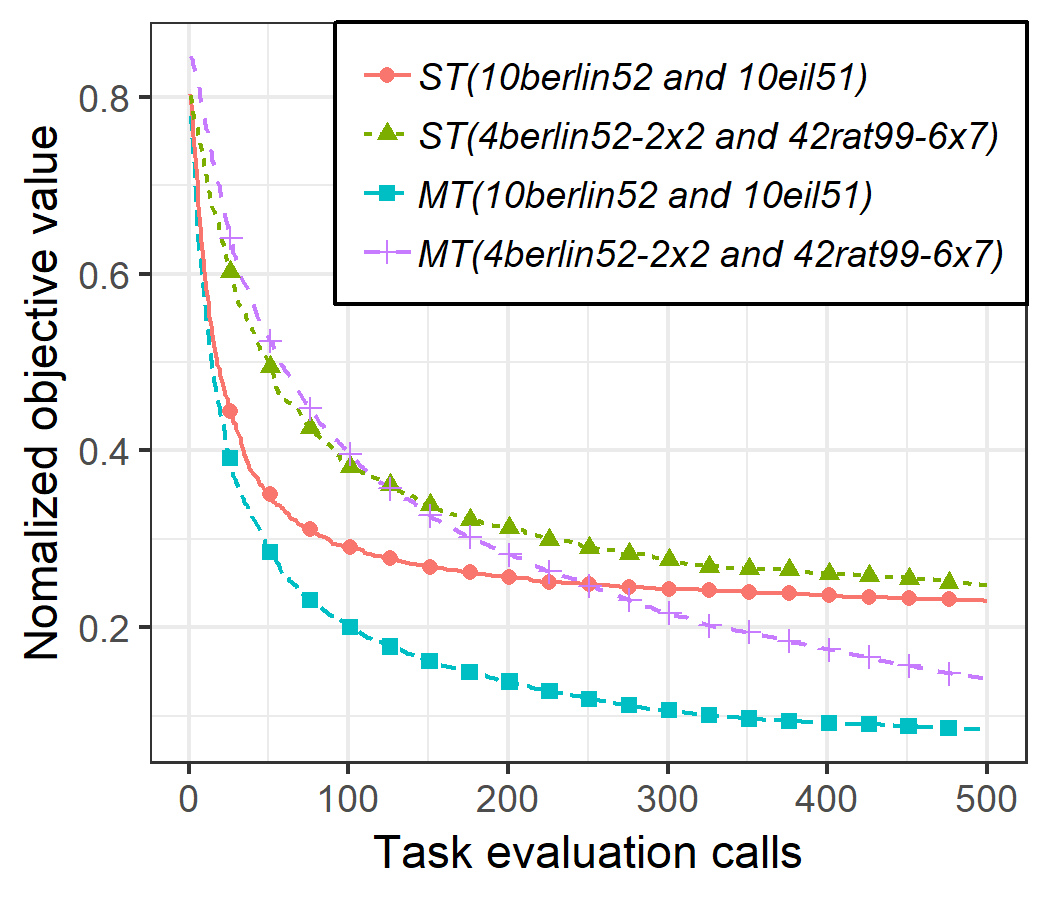}
		\caption{}
		\label{fig:Convergence_Trends}
	\end{subfigure}   
	\begin{subfigure}[b]{.32\linewidth}
		\centering
		\includegraphics[scale=\scalefigure]{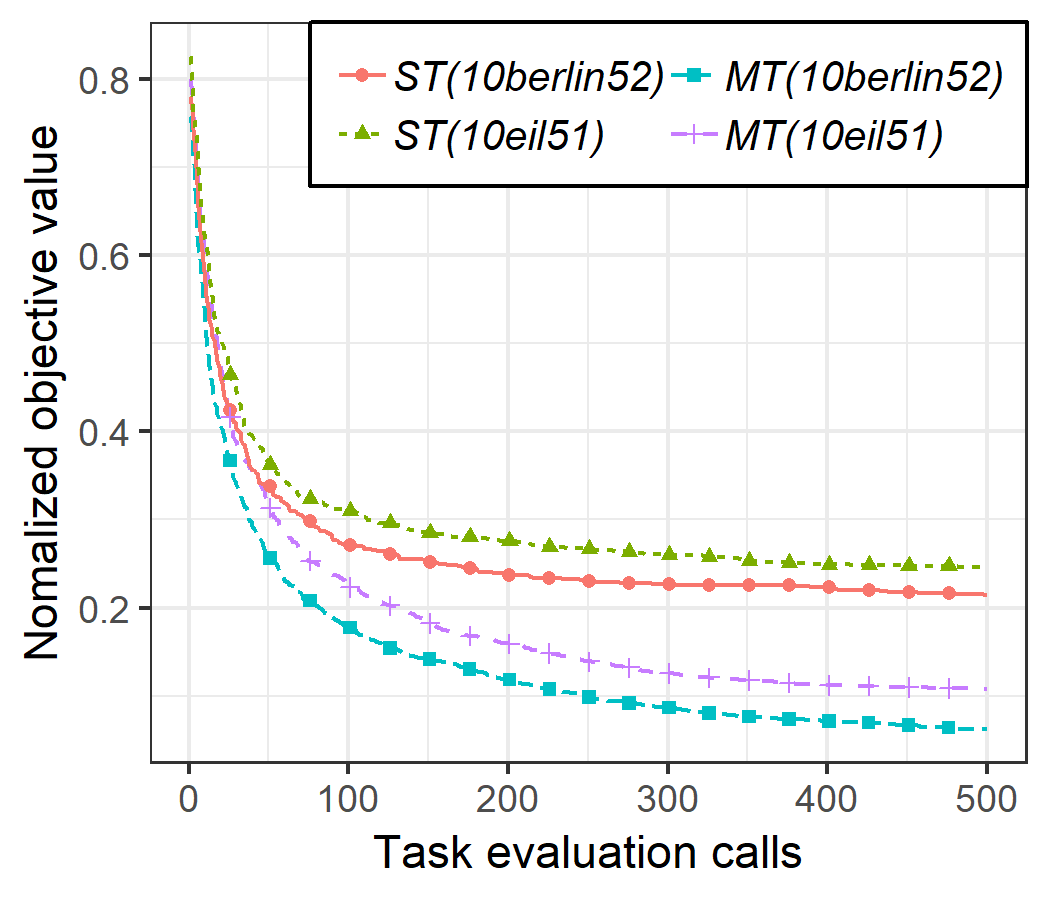}
		\caption{}
		\label{fig:Type_1_Convergence_Each_Task}
	\end{subfigure}
	\begin{subfigure}[b]{.32\linewidth}
		\centering
		\includegraphics[scale=\scalefigure]{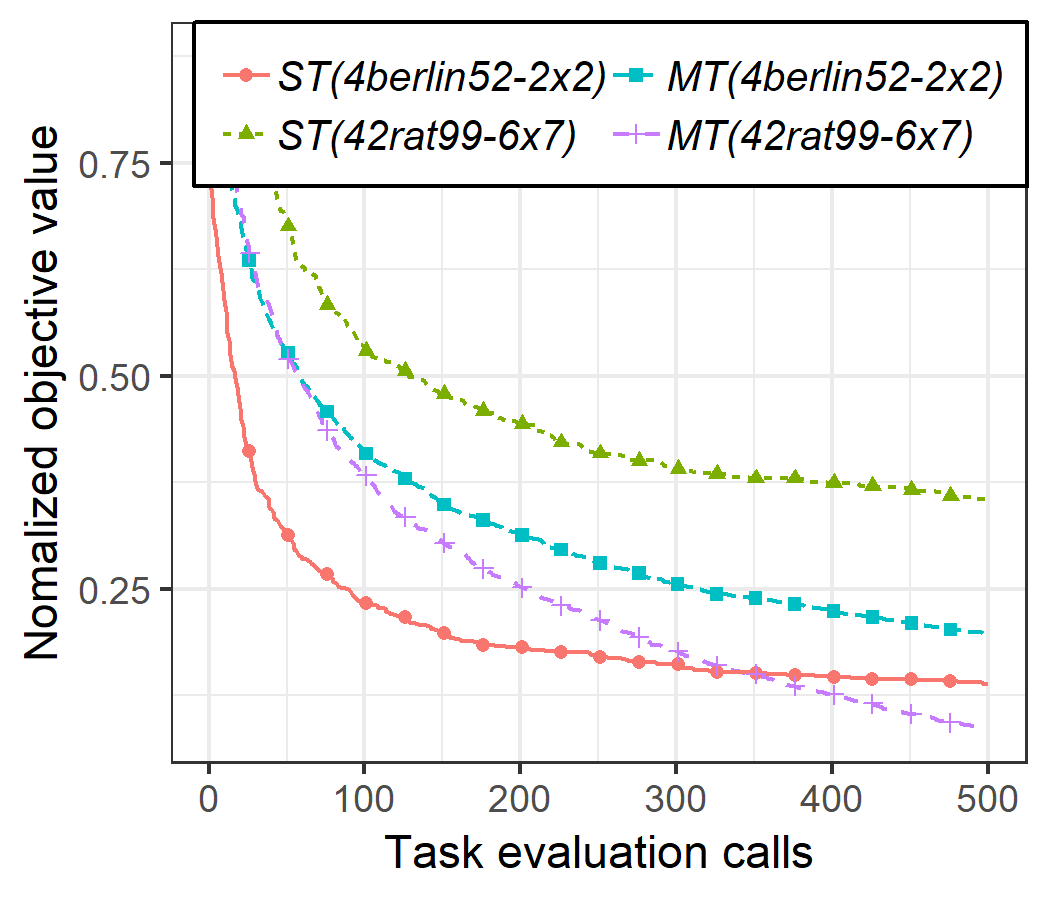}
		\caption{}
		\label{fig:Type_6_Convergence_Each_Task}
	\end{subfigure}
	\caption{Comparing convergence trends of $\tilde{f}$, $\tilde{f_1}$ and $\tilde{f_2}$ in multi-tasks and serial single task for instances in Types 1, 6.}
	\label{fig:ConvergenceTrends}
\end{figure*}

The convergence trend during the initial stages of the multi-tasks is depicted in Figure~\ref{fig:Convergence_Trends} for instances 10berlin52 and 10eil51 in Type 1; instances 4berlin52-2x2 and 42rat99-6x7 in Type 6. In this figure, multi-tasks (MT) converges faster than Single task (ST) at any time in the test instances of Type 1. With Type 6’s instances, ST converges faster than MT at the begin but MT has gradually surpassed ST from evaluation 100 to the end. In summary, the MT performance suggests better overall convergence characteristics than the ST performance.

With the same instances and genetic operators were used in ST and MT, the improvement can be reached because of the exploitation of multiple function landscapes via implicit genetic transfer, such as the evolutionary multitasking paradigm affords.

Refer to Figure~\ref{fig:Type_1_Convergence_Each_Task} and Figure~\ref{fig:Type_6_Convergence_Each_Task}  for a better understanding of the improved performance because of MT. The figure depicts the convergence trends corresponding to each constitutive task. As you can see, in most instances, the convergence of MT is faster in comparison with ST as instances 10berlin52 and 10eil51 in Figure~\ref{fig:Type_1_Convergence_Each_Task}. But, some case, like 4berlin52-2x2 in Figures~\ref{fig:Type_6_Convergence_Each_Task}, the convergence of MT was slower than the one of ST.  In this case, 4berlin52-2x2 and 42rat99-6x7 are simultaneously solved, but two instances are very different structure of solution. 4berlin52-2x2 has 4 clusters and 52 vertices, meanwhile  42rat99-6x7 has 42 clusters and 99 vertices. In unified search space, the shared representation of two solutions is very small, only 4 clusters  42rat99-6x7 impact to 4berlin52-2x2. As a result, the two instances are less impacting each other in process of transferring knowledge during multitasking is the cause of ST outperform than MT.   

\section{Conclusion and discussion}
\label{sec:conclusion}
This paper proposed a multitask optimization algorithm in the realm of the general \gls{mfea} to solve the multiple instances of \gls{clumrct} problem together. Evolutionary operators and a new method for reducing consuming resource of evaluating the solution are also described. The proposed multitasking algorithm is tested on many datasets. Experimental results showed that the proposed \gls{mfea} outperforms its single-task variant in most test cases.


\bibliographystyle{ACM-Reference-Format}
\bibliography{ref}

\onecolumn
\setlength\columnsep{10pt}
\begin{landscape}
\begin{multicols}{2}
	\setbox\ltmcbox\vbox{\makeatletter\col@number\@ne
		\begin{longtable}{ l l r r r r }
\caption{Details of best-found and average fitness per each instance of Type-3 \gls{clumrct} and Type-4 \gls{clumrct} problem solved by \gls{ea} and \gls{mfea}. The best value are highlighted in boldface.}\label{tab:result:type34}\\
\toprule\\

\multicolumn{1}{c}{\textbf{}} & \multicolumn{1}{c}{\textbf{}} & \multicolumn{2}{c}{\textbf{\normalsize{EA}}} & \multicolumn{2}{c}{\textbf{\normalsize{MFEA}}} \\
\cmidrule(l{3pt}r{3pt}){3-4} \cmidrule(l{3pt}r{3pt}){5-6}
& \normalsize{Instances} & \normalsize{BF} & \normalsize{Avg} & \normalsize{BF} & \normalsize{Avg}\\

\midrule
 & 6i300 & 7218321 & \textcolor{black}{7537212.2} & 6892856 & \textbf{\em{7139417.6}}\\

 & 6i350 & 10186959 & \textcolor{black}{10979308.4} & 9528108 & \textbf{\em{10058407.1}}\\

 & 6i400 & 13838525 & \textcolor{black}{14587641.4} & 13052432 & \textbf{\em{14005791.9}}\\

 & 6i450 & 17880773 & \textcolor{black}{19032256.4} & 16420320 & \textbf{\em{17838608.4}}\\

\multirow{-5}{*}{\raggedright\arraybackslash\normalsize \rotatebox{90}{Type 3}} & 6i500 & 23276800 & \textcolor{black}{24761896} & 20823767 & \textbf{\em{22667302.4}}\\
\cmidrule{1-6}
 & 4i200a & 58062080 & \textcolor{black}{64788091.5} & 49873704 & \textbf{\em{57749914.3}}\\

 & 4i200x1 & 35156736 & \textcolor{black}{39884307.3} & 30221411 & \textbf{\em{36285931.8}}\\

 & 4i200z & 35580760 & \textcolor{black}{39847121.9} & 31923515 & \textbf{\em{37291083.8}}\\

\multirow{-4}{*}{\raggedright\arraybackslash\normalsize \rotatebox{90}{Type 4}} & 4i400a & 348050079 & \textcolor{black}{381840774.9} & 235345395 & \textbf{\em{273783937.4}}\\

 & 4i400h & 203858230 & \textcolor{black}{224087266.5} & 179831484 & \textbf{\em{204974618.1}}\\

 & 4i400x1 & 204931109 & \textcolor{black}{223854804.4} & 178575229 & \textbf{\em{202321240.4}}\\

 & 4i400x2 & 211600994 & \textcolor{black}{224658063.9} & 184342319 & \textbf{\em{204305625.5}}\\

 & 4i400z & 189307036 & \textcolor{black}{221169556.9} & 174142050 & \textbf{\em{202083877.8}}\\
\bottomrule
\end{longtable}

\unskip\unpenalty\unpenalty
	}\unvbox\ltmcbox
	
	\setbox\ltmcbox\vbox{\makeatletter\col@number\@ne
		\begin{longtable}{ l l r r r r }
	\caption{Details of best-found and average fitness per each instance of Type-5 \gls{clumrct} problem solved by \gls{ea} and \gls{mfea}. The best value are highlighted in boldface.}\label{tab:result:type5}\\
	\toprule\\
	
	\multicolumn{1}{c}{\textbf{}} & \multicolumn{1}{c}{\textbf{}} & \multicolumn{2}{c}{\textbf{\normalsize{EA}}} & \multicolumn{2}{c}{\textbf{\normalsize{MFEA}}} \\
	\cmidrule(l{3pt}r{3pt}){3-4} \cmidrule(l{3pt}r{3pt}){5-6}
	& \normalsize{Instances} & \normalsize{BF} & \normalsize{Avg} & \normalsize{BF} & \normalsize{Avg}\\
	
	\midrule
	& 10i120-46 & 8620633.2 & \textcolor{black}{9393459.8} & 7784832.4 & \textbf{\em{8197268.9}}\\
	
	& 10i45-18 & 875824.2 & \textcolor{black}{932734.6} & 756198.1 & \textbf{\em{851733.6}}\\
	
	& 10i60-21 & 1798836.1 & \textcolor{black}{2069872.8} & 1598921.1 & \textbf{\em{1853126.2}}\\
	
	& 10i65-21 & 2379511.4 & \textcolor{black}{2491125.7} & 1996599.7 & \textbf{\em{2139469.2}}\\
	
	& 10i70-21 & 2630883.1 & \textcolor{black}{2873851.8} & 2324482.7 & \textbf{\em{2546084.3}}\\
	
	& 10i90-33 & 3889081.6 & \textcolor{black}{4150252.1} & 3135916.2 & \textbf{\em{3504309.9}}\\
	
	& 5i120-46 & 7549373.0 & \textcolor{black}{8134562.4} & 6244382.8 & \textbf{\em{6933990.9}}\\
	
	& 5i30-17 & 340097.8 & \textbf{\em{358611.6}} & 343481.4 & \textcolor{black}{368298.3}\\
	
	& 5i45-18 & 750733.1 & \textcolor{black}{802806.8} & 658041.4 & \textbf{\em{714002.8}}\\
	
	& 5i60-21 & 1771839.1 & \textcolor{black}{1974753.5} & 1581756.3 & \textbf{\em{1746774.6}}\\
	
	& 5i65-21 & 2132123.3 & \textcolor{black}{2311904.2} & 1948779.0 & \textbf{\em{2142993.9}}\\
	
	& 5i70-21 & 2580773.5 & \textcolor{black}{2749384.8} & 2135118.7 & \textbf{\em{2410433.1}}\\
	
	& 5i75-22 & 2740946.6 & \textcolor{black}{3012711} & 2527119.8 & \textbf{\em{2760425.4}}\\
	
	& 5i90-33 & 3604939.2 & \textcolor{black}{3971655.3} & 3187292.2 & \textbf{\em{3407267.2}}\\
	
	& 7i60-21 & 1960693.7 & \textcolor{black}{2139328.6} & 1637152.6 & \textbf{\em{1848278.6}}\\
	
	\multirow{-16}{*}{\raggedright\arraybackslash\normalsize \rotatebox{90}{Small Instances}} & 7i65-21 & 2266953.8 & \textcolor{black}{2475010.1} & 1917674.5 & \textbf{\em{2121462}}\\
	\cmidrule{1-6}
	& 10i300-109 & 47618138.3 & \textcolor{black}{49995372.1} & 40551819.0 & \textbf{\em{45274502.3}}\\
	
	& 10i400-206 & 108421226.4 & \textcolor{black}{115141165.4} & 88310142.6 & \textbf{\em{99012666}}\\
	
	& 10i500-305 & 182614330.7 & \textcolor{black}{198554467.7} & 138313818.7 & \textbf{\em{145731966.4}}\\
	
	& 15i300-110 & 49192926.5 & \textcolor{black}{54123296.2} & 43843232.5 & \textbf{\em{47318804.9}}\\
	
	& 15i400-207 & 102953554.8 & \textcolor{black}{110503351.8} & 95421482.9 & \textbf{\em{103134953.9}}\\
	
	& 15i500-306 & 176477520.4 & \textcolor{black}{185003609.4} & 156507079.7 & \textbf{\em{164366536.1}}\\
	
	& 20i300-111 & 60901743.1 & \textcolor{black}{64793312.9} & 53572569.6 & \textbf{\em{59089047.3}}\\
	
	& 20i400-208 & 112865091.8 & \textcolor{black}{120286857} & 99758157.6 & \textbf{\em{105914056.3}}\\
	
	& 20i500-307 & 157448014.7 & \textcolor{black}{164840517.5} & 114601072.8 & \textbf{\em{122660401.5}}\\
	
	& 25i300-112 & 57643197.6 & \textcolor{black}{61742234} & 49921091.2 & \textbf{\em{52990412.1}}\\
	
	& 25i400-209 & 115888694.2 & \textcolor{black}{123273581.5} & 101778949.6 & \textbf{\em{110739520.8}}\\
	
	& 25i500-308 & 174495737.9 & \textcolor{black}{186917007.2} & 148853209.8 & \textbf{\em{162658959.2}}\\
	
	& 5i300-108 & 59885205.6 & \textcolor{black}{64875737.1} & 56732257.9 & \textbf{\em{61339474.6}}\\
	
	\multirow{-14}{*}{\raggedright\arraybackslash\normalsize \rotatebox{90}{Large instances}} & 5i400-205 & 86596796.4 & \textcolor{black}{92361135.8} & 72401881.2 & \textbf{\em{79314034.6}}\\
	\bottomrule
\end{longtable}
\unskip\unpenalty\unpenalty
		
	}\unvbox\ltmcbox
	
		\setbox\ltmcbox\vbox{\makeatletter\col@number\@ne
		\begin{longtable}{ l l r r r r }
\caption{Details of best-found and average fitness per each instance of Type-6 \gls{clumrct} problem solved by \gls{ea} and \gls{mfea}. The best value are highlighted in boldface.}\label{tab:result:type6}\\
\toprule\\

\multicolumn{1}{c}{\textbf{}} & 
\multicolumn{1}{c}{\textbf{}} & 
\multicolumn{2}{c}{\textbf{\normalsize{EA}}} & 
\multicolumn{2}{c}{\textbf{\normalsize{MFEA}}} \\
\cmidrule(l{3pt}r{3pt}){3-4} \cmidrule(l{3pt}r{3pt}){5-6}

& \normalsize{Instances} & \normalsize{BF} & \normalsize{Avg} & \normalsize{BF} & \normalsize{Avg}\\

\midrule
 & 10berlin52-2x5 & 1265672.9 & \textcolor{black}{1364770.7} & 1164646.4 & \textbf{\em{1235853.7}}\\

 & 12eil51-3x4 & 70786.4 & \textcolor{black}{74726.4} & 66650.9 & \textbf{\em{72092.9}}\\

 & 12eil76-3x4 & 183970.9 & \textcolor{black}{197678.8} & 159666.4 & \textbf{\em{171642.9}}\\

 & 12pr76-3x4 & 40150030.5 & \textcolor{black}{42816754.4} & 35017437.6 & \textbf{\em{37099269.8}}\\

 & 12st70-3x4 & 222752.5 & \textcolor{black}{235800.4} & 200185.7 & \textbf{\em{211882.1}}\\

 & 15pr76-3x5 & 39342027.9 & \textcolor{black}{43671902.5} & 36084874.9 & \textbf{\em{40125991.8}}\\

 & 16eil51-4x4 & 68868.2 & \textcolor{black}{73854.8} & 61079.9 & \textbf{\em{67377.2}}\\

\multirow{-7}{*}{\raggedright\arraybackslash\normalsize \rotatebox{90}{Small Instances}} 

 & 16eil76-4x4 & 180451.0 & \textcolor{black}{195833.1} & 161420.6 & \textbf{\em{170210}}\\

 & 16lin105-4x4 & 12404567.1 & \textcolor{black}{12957738.6} & 10755239.4 & \textbf{\em{11622527.4}}\\

 & 16st70-4x4 & 217573.7 & \textcolor{black}{234338.3} & 198685.6 & \textbf{\em{214505.5}}\\

 & 18pr76-3x6 & 41023837.1 & \textcolor{black}{43177318.2} & 34103588.2 & \textbf{\em{35339402.2}}\\

 & 20eil51-4x5 & 72362.6 & \textcolor{black}{76595.1} & 65334.0 & \textbf{\em{69527.5}}\\

 & 20eil76-4x5 & 178006.8 & \textcolor{black}{193912.2} & 158730.3 & \textbf{\em{170962.8}}\\

 & 20st70-4x5 & 224071.9 & \textcolor{black}{240294.9} & 197197.6 & \textbf{\em{218051.4}}\\

 & 25eil101-5x5 & 363706.0 & \textcolor{black}{404274.6} & 326204.4 & \textbf{\em{353615.7}}\\

 & 25eil51-5x5 & 76000.5 & \textcolor{black}{80150.8} & 64662.7 & \textbf{\em{70299.9}}\\

 & 25eil76-5x5 & 185727.3 & \textcolor{black}{201911.2} & 152927.8 & \textbf{\em{176593.9}}\\

 & 25rat99-5x5 & 775778.4 & \textcolor{black}{842529.3} & 672730.4 & \textbf{\em{734618.1}}\\

 & 28kroA100-4x7 & 16594234.3 & \textcolor{black}{18057077.9} & 13127755.6 & \textbf{\em{15332145.5}}\\

 & 30kroB100-5x6 & 16197360.5 & \textcolor{black}{17777465.4} & 13321104.8 & \textbf{\em{15180119.9}}\\

 & 35kroB100-5x5 & 15975455.4 & \textcolor{black}{17175668.5} & 13207171.0 & \textbf{\em{14039628.2}}\\

 & 36eil101-6x6 & 400479.2 & \textcolor{black}{444799.2} & 301954.6 & \textbf{\em{336621.4}}\\

 & 42rat99-6x7 & 957728.7 & \textcolor{black}{1036331.1} & 726586.0 & \textbf{\em{801111.5}}\\

 & 4berlin52-2x2 & 1204074.6 & \textbf{\em{1389348.5}} & 1407569.0 & \textcolor{black}{1468442.6}\\

 & 4eil51-2x2 & 69289.7 & \textcolor{black}{79233.1} & 63352.3 & \textbf{\em{74463.2}}\\

 & 4eil76-2x2 & 193695.1 & \textcolor{black}{216040.9} & 160470.3 & \textbf{\em{173950.9}}\\

 & 4pr76-2x2 & 43149835.2 & \textcolor{black}{48937295.2} & 37161565.6 & \textbf{\em{39648471.3}}\\

 & 6berlin52-2x3 & 1250239.7 & \textcolor{black}{1387090.3} & 1211907.4 & \textbf{\em{1338852.9}}\\

 & 6pr76-2x3 & 39811089.4 & \textcolor{black}{44505653.2} & 38034663.2 & \textbf{\em{40323875.3}}\\

 & 6st70-2x3 & 244346.0 & \textcolor{black}{262236.4} & 210941.4 & \textbf{\em{236206.3}}\\

 & 8berlin52-2x4 & 1335897.9 & \textcolor{black}{1412954.5} & 1161547.7 & \textbf{\em{1248068.8}}\\

 & 9eil101-3x3 & 382895.4 & \textcolor{black}{408022.9} & 323399.9 & \textbf{\em{347184.1}}\\

 & 9eil51-3x3 & 73735.2 & \textcolor{black}{78254.5} & 65117.7 & \textbf{\em{69709.3}}\\

 & 9eil76-3x3 & 185572.5 & \textcolor{black}{200795.4} & 159175.3 & \textbf{\em{171531}}\\
\cmidrule{1-6}
 & 18pr439-3x6 & 1416839349.6 & \textcolor{black}{1561793571.1} & 1025085049.5 & \textbf{\em{1137123264.7}}\\

 & 20pr439-4x5 & 1477471558.3 & \textcolor{black}{1613475583} & 1305604456.7 & \textbf{\em{1485293184.2}}\\

 & 25a280-5x5 & 12349778.6 & \textcolor{black}{13110863.9} & 10585336.1 & \textbf{\em{11673178.5}}\\

 & 25gil262-5x5 & 8614336.6 & \textcolor{black}{9381304.6} & 7589106.3 & \textbf{\em{8232173.4}}\\

 & 25pcb442-5x5 & 488506236.5 & \textcolor{black}{530402249.8} & 406965163.9 & \textbf{\em{446099393.2}}\\

 & 36pcb442-6x6 & 508715616.5 & \textcolor{black}{553060407.9} & 354212969.5 & \textbf{\em{408803078}}\\

 & 42a280-6x7 & 12137547.4 & \textcolor{black}{14334181.3} & 9689989.9 & \textbf{\em{10779328.7}}\\

 & 49gil262-7x7 & 10277303.1 & \textcolor{black}{11437109.1} & 8263400.4 & \textbf{\em{9904810.3}}\\

 & 49lin318-7x7 & 270088847.1 & \textcolor{black}{309400144.5} & 221116828.4 & \textbf{\em{257834438.5}}\\

 & 9a280-3x3 & 14848459.2 & \textcolor{black}{15930228.1} & 12543766.2 & \textbf{\em{14175273.4}}\\

 & 9gil262-3x3 & 10732649.9 & \textcolor{black}{11851918.2} & 9641118.4 & \textbf{\em{10818917.3}}\\

 & 9lin318-3x3 & 293171349.6 & \textcolor{black}{314380349} & 269984382.8 & \textbf{\em{293792191.7}}\\

\multirow{-13}{*}{\raggedright\arraybackslash\normalsize \rotatebox{90}{Large instances}} & 9pcb442-3x3 & 657120275.1 & \textcolor{black}{706429043.6} & 541118426.3 & \textbf{\em{579184857.7}}\\
\bottomrule
\end{longtable}
\unskip\unpenalty\unpenalty
	}\unvbox\ltmcbox
		\setbox\ltmcbox\vbox{\makeatletter\col@number\@ne
		\begin{longtable}{ l l r r r r }
\caption{Details of best-found and average fitness per each instance of Type-1 CluMRCT problem solved by EA and MFEA. The best value are highlighted in boldface.}\label{tab:result:type1}\\
\toprule\\

\multicolumn{1}{c}{\textbf{}} & \multicolumn{1}{c}{\textbf{}} & \multicolumn{2}{c}{\textbf{\normalsize{EA}}} & \multicolumn{2}{c}{\textbf{\normalsize{MFEA}}} \\
\cmidrule(l{3pt}r{3pt}){3-4} \cmidrule(l{3pt}r{3pt}){5-6}
 & \normalsize{Instances} & \normalsize{BF} & \normalsize{Avg} & \normalsize{BF} & \normalsize{Avg}\\
 
\midrule
 & 10berlin52 & 1212367.7 & \textcolor{black}{1270287.3} & 1070892.2 & \textbf{\em{1128537.5}}\\

 & 10eil51 & 72400.6 & \textcolor{black}{75851} & 62220.9 & \textbf{\em{68196.7}}\\

 & 10eil76 & 182260.6 & \textcolor{black}{197520.7} & 154720.2 & \textbf{\em{176318.9}}\\

 & 10kroB100 & 15544670.3 & \textcolor{black}{16793258.9} & 13410213.4 & \textbf{\em{14430156.1}}\\

 & 10pr76 & 39803806.2 & \textcolor{black}{41978071.8} & 34443906.0 & \textbf{\em{37528661.7}}\\

 & 10rat99 & 787308.7 & \textcolor{black}{844359.2} & 681410.1 & \textbf{\em{733878.6}}\\

 & 15eil51 & 71494.2 & \textcolor{black}{75430.9} & 65536.1 & \textbf{\em{69262.8}}\\

 & 15eil76 & 171846.9 & \textcolor{black}{190699.4} & 152276.6 & \textbf{\em{165567.5}}\\

 & 15pr76 & 38723411.2 & \textcolor{black}{41355082} & 34406443.4 & \textbf{\em{35902210.1}}\\

 & 15st70 & 217633.3 & \textcolor{black}{229835} & 196100.4 & \textbf{\em{213158.6}}\\

 & 25eil101 & 334091.7 & \textcolor{black}{382588.2} & 298566.2 & \textbf{\em{331092.9}}\\

 & 25kroA100 & 15122586.0 & \textcolor{black}{16750515.8} & 13290236.0 & \textbf{\em{14801709.7}}\\

 & 25lin105 & 10982060.6 & \textcolor{black}{11914969.5} & 155233.0 & \textbf{\em{281917}}\\

 & 25rat99 & 784286.8 & \textbf{\em{833241.1}} & 638380.1 & \textcolor{black}{968377.2}\\

 & 50eil101 & 479713.2 & \textcolor{black}{514738.4} & 9788.0 & \textbf{\em{12654}}\\

 & 50kroA100 & 21002827.6 & \textbf{\em{22851798.3}} & 22757582.7 & \textcolor{black}{30014740.7}\\

 & 50kroB100 & 19047316.1 & \textcolor{black}{22503440.4} & 491024.2 & \textbf{\em{606069.8}}\\

 & 50lin105 & 15309391.6 & \textbf{\em{17006431.4}} & 14536077.1 & \textcolor{black}{23186711.8}\\

 & 50rat99 & 997133.3 & \textbf{\em{1178858.6}} & 1040326.9 & \textcolor{black}{1404530}\\

 & 5berlin52 & 1251233.9 & \textcolor{black}{1360740} & 27351.5 & \textbf{\em{32335.8}}\\

 & 5eil51 & 73451.1 & \textcolor{black}{79603.6} & 1955.0 & \textbf{\em{2198.5}}\\

 & 5eil76 & 198578.3 & \textcolor{black}{211390.5} & 129445.0 & \textbf{\em{143326.5}}\\

 & 5pr76 & 45587589.8 & \textcolor{black}{48401813.1} & 667837.8 & \textbf{\em{822934.7}}\\

\multirow{-24}{*}{\raggedright\arraybackslash\normalsize \rotatebox{90}{Small Instances}} & 5st70 & 240408.6 & \textcolor{black}{260969.9} & 168499.6 & \textbf{\em{187038.6}}\\
\cmidrule{1-6}
 & 10a280 & 13907739.0 & \textcolor{black}{14844887.5} & 12300290.1 & \textbf{\em{13350381.4}}\\

 & 10gil262 & 10583888.3 & \textcolor{black}{11299878.5} & 9267931.7 & \textbf{\em{10283586.5}}\\

 & 10lin318 & 274950582.8 & \textcolor{black}{297714185.9} & 237759185.5 & \textbf{\em{270759068.6}}\\

 & 10pcb442 & 623457163.9 & \textcolor{black}{685302098.3} & 505781275.6 & \textbf{\em{547646940.6}}\\

 & 10pr439 & 1390944006.2 & \textcolor{black}{1502442441.3} & 855937241.5 & \textbf{\em{895455630.7}}\\

 & 25a280 & 10920850.4 & \textcolor{black}{11996040.1} & 9761001.7 & \textbf{\em{10244430.5}}\\

 & 25gil262 & 8320804.1 & \textcolor{black}{9416167.1} & 7615763.2 & \textbf{\em{8364538.5}}\\

 & 25lin318 & 225383682.6 & \textcolor{black}{251311451.3} & 195768659.3 & \textbf{\em{211302877.7}}\\

 & 25pcb442 & 476613112.1 & \textcolor{black}{509512270.5} & 412614081.5 & \textbf{\em{451957890.3}}\\

\multirow{-10}{*}{\raggedright\arraybackslash\normalsize \rotatebox{90}{Large instances}} & 25pr439 & 1162429357.3 & \textcolor{black}{1213987542.9} & 955864371.0 & \textbf{\em{1022805770.1}}\\
\bottomrule
\end{longtable}

\unskip\unpenalty\unpenalty
	}\unvbox\ltmcbox
\end{multicols}
\end{landscape}

%
%
%
%

\end{document}